\pdfoutput=1
\documentclass[11pt]{article}

\usepackage{acl}

\usepackage{times}
\usepackage{latexsym}
\usepackage{multirow}
\usepackage{CJKutf8}

\usepackage{tabularx}
\usepackage{float}
\usepackage{graphicx}
\usepackage{subfigure}
\usepackage{caption}
\usepackage{booktabs}
\usepackage{longtable}
\usepackage{supertabular}
\usepackage{arydshln} 
\usepackage{booktabs}
\usepackage{rotating}
\usepackage{url}
\usepackage{enumitem}
\setenumerate[1]{itemsep=0pt,partopsep=0pt,parsep=\parskip,topsep=5pt}
\setitemize[1]{itemsep=0pt,partopsep=0pt,parsep=\parskip,topsep=5pt}
\setdescription{itemsep=0pt,partopsep=0pt,parsep=\parskip,topsep=5pt}

\usepackage[T1]{fontenc}
\usepackage[utf8]{inputenc}

\usepackage{microtype}

\title{CSL: A Large-scale Chinese Scientific Literature Dataset}

\author{Yudong Li$^{1,2}$, Yuqing Zhang$^{1*}$, Zhe Zhao$^{3}$, Linlin Shen$^{2}$, Weijie Liu$^{3}$, \\ 
{\bf  Weiquan Mao$^{3}$, and Hui Zhang$^{4}$} \\
$^1$ China University of Geosciences (Beijing), School of Information Engineering \\
$^2$ Shenzhen University, School of Computer Science and Software Engineering \\
$^3$ Tencent AI Lab\\
$^4$ Information Technology Center for National Science \\ and Technology Infrastructure, Beijing, China
}

\begin{document}
\maketitle
\begin{abstract}
Scientific literature serves as a high-quality corpus, supporting a lot of Natural Language Processing (NLP) research.
However, existing datasets are centered around the English language, which restricts the development of Chinese scientific NLP.
In this work, we present CSL, a large-scale \textbf{C}hinese \textbf{S}cientific \textbf{L}iterature dataset, which contains the titles, abstracts, keywords and academic fields of 396k papers. To our knowledge, CSL is the first scientific document dataset in Chinese. % and provides discipline annotation.
The CSL can serve as a Chinese corpus. Also, this semi-structured data is a natural annotation that can constitute many supervised NLP tasks. Based on CSL, we present a benchmark to evaluate the performance of models across scientific domain tasks, i.e., summarization, keyword generation and text classification. 
We analyze the behavior of existing text-to-text models on the evaluation tasks and reveal the challenges for Chinese scientific NLP tasks, which provides a valuable reference for future research. Data and code are available at \url{https://github.com/ydli-ai/CSL} .
\end{abstract}

% The rich semantic information brings extensive NLP tasks, based on which a benchmark is introduced to evaluate the performance of models across scientific domain tasks, e.g. summarization and keyword generation.

\section{Introduction}

\begin{figure}[t]
\centerline{\includegraphics[width=8cm]{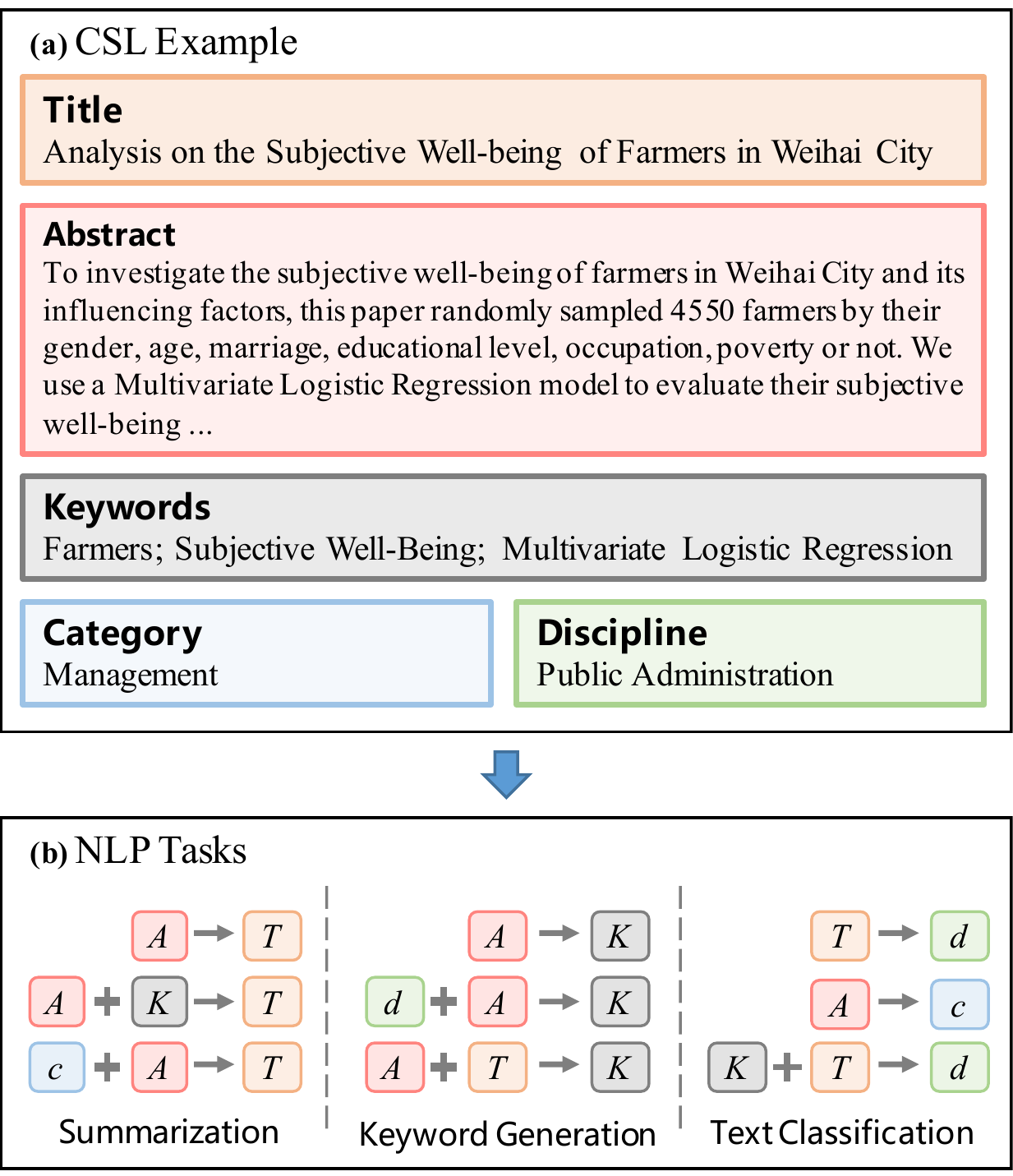}}
\caption{(a) An example of paper meta-information (translated into English). (b) Examples of NLP tasks constructed from CSL. The arrow indicates the input and output of the task, for example, ``\textit{A $\to$ T}'' represents the task feeding abstract to produce title. \textit{A}: abstract; \textit{T}: title; \textit{K}: keywords; \textit{c}: category; \textit{d}: discipline.}
\label{f1}
%\vspace{-0.5cm}
\end{figure}

\begin{table*}[t]
\small
\centering
%\resizebox{1.00\columnwidth}{!}{
\begin{tabular}{lrrrrrl}
\toprule
\textbf{Category} & \#\textit{d} & len(T) & len(A) & num(K) & \#Samples & Discipline Examples \\ \midrule
\textbf{Engineering} & 27 & 19.1 & 210.9 & 4.4 & 177,600 & \textit{Mechanics, Architecture, Electrical Science}  \\
\textbf{Science} & 9 & 20.7 & 254.4 & 4.3 & 35,766 & \textit{Mathematics, Physics, Astronomy, Geography} \\
\textbf{Agriculture} & 7 & 17.1 & 177.1 & 7.1 & 39,560 & \textit{Crop Science, Horticulture, Forestry} \\
\textbf{Medicine} & 5 & 20.7 & 269.5 & 4.7 & 36,783 & \textit{Clinical Medicine, Dental Medicine, Pharmacy} \\
\textbf{Management} & 4 & 18.7 & 157.7 & 6.2 & 23,630 & \textit{Business Management, Public Administration} \\
\textbf{Jurisprudence} & 4 & 18.9 & 174.4 & 6.1 & 21,554 & \textit{Legal Science, Political Science, Sociology} \\
\textbf{Pedagogy} & 3 & 17.7 & 179.4 & 4.3 & 16,720 & \textit{Pedagogy, Psychology, Physical Education} \\
\textbf{Economics} & 2 & 19.5 & 177.2 & 4.5 & 11,558 & \textit{Theoretical Economics, Applied Economics} \\
\textbf{Literature} & 2 & 18.8 & 158.2 & 8.3 & 10,501 & \textit{Chinese Literature, Journalism} \\
\textbf{Art} & 1 & 17.8 & 170.8 & 5.4 & 5,201 & \textit{Art} \\
\textbf{History} & 1 & 17.6 & 181.0 & 6.0 & 6,270 & \textit{History} \\
\textbf{Strategics} & 1 & 17.5 & 169.3 & 4.0 & 3,555 & \textit{Military Science} \\
\textbf{Philosophy} & 1 & 18.0 & 176.5 & 8.0 & 7,511 & \textit{Philosophy} \\ \midrule %\cdashline{1-7}[1pt/1pt]
\textbf{All} & 67 &  &  &  & 396,209 &  \\ \bottomrule
\end{tabular}
%}
\caption{\label{statistics}
Detailed statistics of the CSL dataset. \#d: The number of disciplines in the category. len(T): Average length of each title; len(A): Average length of each abstract; num(K): Average number of keywords. 
}
%\vspace{-0.5cm}
\end{table*}

With the increase in the publication of papers, Natural Language Processing (NLP) tools that assist users in writing, searching, and archiving scientific literature have grown increasingly important.
For instance, paper/citation recommendation \cite{beel2016paper, cohan2020specter}, topic classification \cite{beltagy2019scibert} and summarization \cite{cohan2018scientific} systems have been developed. 
The construction of these automatic systems primarily relies on academic resources such as large-scale corpus \citep{lo2020s2orc, saier2020unarxive}, citation graphs \citep{sinha2015overview, tang2008arnetminer, zhang2019oag} and supervised scientific datasets \cite{li2016biocreative, jurgens2018measuring}.
These resources, however, are mostly centered around the English language, which restricts the development of techniques for addressing non-English scientific NLP tasks. 
Until recently, progresses in the NLP research for Chinese resources and models has lagged behind their English counterparts.

To fill the gap of non-English scientific resources and spur the Chinese scientific NLP research, in this paper, we introduce CSL: a large-scale \textbf{C}hinese \textbf{S}cientific \textbf{L}iterature dataset.
CSL contains 396,209 Chinese papers' meta-information, including title, abstract, keywords, academic category and discipline. 
Papers are collected from comprehensive Chinese academic journals covering a wide range of distribution.
In particular, we divide them into 13 first-level categories and 67 second-level disciplines. 
%Compared to existing academic resources \cite{lo2020s2orc,saier2019bibliometric}, we provide broader and more fine-grained research fields.
In addition to the difference in language, it provides broader and more fine-grained research fields than existing academic resources \cite{lo2020s2orc,saier2019bibliometric}.

%CSL can serve as a pre-training corpus and can be used for data mining.
Scientific literature metadata contains abundant semantic information, making it a natural annotated data source with the potential to provide many NLP tasks. 
For example, predicting the title with abstract constitutes a summarization task. As the data and task examples shown in Figure~\ref{f1}, such combinations can constitute abundant tasks.
These tasks can drive models in real-world applications and are essential for a lot of academic NLP research.
To better understand the challenges posed by Chinese scientific NLP, we build a benchmark consisting of a series of CSL-derived tasks, i.e., summarization, keyword generation and category/discipline classification. We also provide a toolkit that allows users to design evaluation tasks according to their needs.

We implement some state-of-the-art Chinese text-to-text models and evaluate on the proposed benchmark. 
We also demonstrate the effectiveness of the CSL dataset as pre-training corpus. Specifically, we pre-train T5 with paper abstracts, namely CSL-T5. It outperforms the model trained on the general-domain corpus, which can be used as a strong baseline for the proposed benchmark.
The experiment results show that though existing models can achieve acceptable performance on scientific NLP tasks, it still needs future efforts to reach a practical level.

The main contributions of this paper are summarized as follows:

\begin{itemize}

\item  We release the first large-scale Chinese Scientific Literature dataset (CSL), which can be used for many different purposes, e.g., pre-training corpus and scientific-related tasks.

\item  Based on the CSL, we build a benchmark that represents real-world scenarios of automatic analyzing scientific literature.

\item  We implement text-to-text models to provide baselines. The experimental results highlight the model's difficulties in Chinese scientific NLP tasks.

\end{itemize}

\section{The CSL Dataset}

\begin{table*}[]
\small
\centering
\begin{tabular}{llllll}
\toprule
Dataset & Instances & Language & Peer Review & Source & Academic Disciplines \\ \midrule
S2ORC \citeyearpar{lo2020s2orc} & 8.1M & English & not all & MAG, arXiv, PubMed & 20 (multi) \\
PubMed Central OAS & 2.3M & English & not all &  PubMed & 2 (bio, LS) \\
unarXive \citeyearpar{saier2020unarxive} & 1.0M & English & not all & MAG, arXiv & 4 (physics, math, CS, other) \\ 
\citealp{saier2019bibliometric} & 1.0M & English & not all & arXiv & 3 (physics, math, CS) \\
arXiv CS \citeyearpar{farber2018high} & 90k & English & not all & arXiv & 1 (CS) \\
AAN \citeyearpar{radev2013acl} & 25k & English & all & ACL Anthology & 1 (comp ling) \\
\midrule
CSL (ours) & 396k & Chinese & all & Chinese Core Journals &  67 (multi) \\ \bottomrule
\end{tabular}
\caption{\label{comparison}
A comparison of CSL with other publicly-available scientific literature datasets. Note that we provide the first dataset in Chinese, which also has the more fine-grained discipline annotation. bio: biomedicine; LS: life science; CS: computer science; comp ling: computational linguistics. 
}
\end{table*}

\subsection{Data Collection}

We obtain the paper's meta-information from the National Engineering Research Center for Science and Technology Resources Sharing Service (NSTR) \footnote{https://nstr.escience.net.cn} dated from 2010 to 2020. 
Then, we filter data by the Catalogue of Chinese Core Journals, which is an academic journal evaluation standard published by Peking University Library. It selects 1,980 core journals from the Chinese journals, widely recognized by the Chinese academic community.

According to the Catalogue and collected data, we divide academic fields into 13 first-level categories (e.g., Engineering, Science) and 67 second-level disciplines (e.g., Mechanics, Mathematics).
We use the journal's instructions to assign journals to categories and disciplines, and only journals that focus on a single academic field are kept.
For the guideline of human annotation, we follow the Disciplines of Conferring Academic Degrees (GB/T 13745-2009). We ask volunteers to read the introduction of the journal and find the closest discipline from the guideline.
As a result, papers can be labeled with categories and disciplines based on the journal in which they were published.
For example, papers from the ``Chinese Journal of Computers'' are categorized into the category ``Engineering'' and the discipline ``Computer Science''.

In total, we collect 396,209 instances for the CSL dataset, represented as tuples $<T,A,K,c,d>$, where $T$ is the title, $A$ is the abstract, $K$ is a list of keywords, $c$ is the category label and $d$ is the discipline label.
Due to the ethical concern, we only use the paper's publicly available meta-information and do not access the full text.

\subsection{Data Analysis}

The paper distribution over categories and the examples of disciplines are shown in Table~\ref{statistics}.
%Among the categories, Engineering contains the most disciplines, therefore has the largest sample size at 44.8\%.
%Some categories contain only one discipline, which is difficult to further divide due to the limited samples we have collected.
A total of 67 disciplines are collected by CSL, covering almost all research fields. Each discipline contains 3000-10000 samples.

Table~\ref{comparison} presents an overview of existing academic datasets.
In comparison, the CSL dataset has the following features: 
\textbf{(1) Wider discipline coverage.} 
Existing datasets mainly focus on specific domains, while CSL covers almost all research domains. Also, CSL has more fine-grained discipline labels.
\textbf{(2) New data source.} 
It can be seen that existing datasets are largely built on digital libraries like arXiv \footnote{https://arxiv.org}, PubMed \footnote{http://www.pubmed.gov}, ACL Anthology \footnote{https://aclanthology.org} and MAG \cite{sinha2015overview}, resulting in some overlap.
CSL presents a new data source in Chinese that complements existing resources.
\textbf{(3) Higher quality and accuracy.} 
Digital libraries contain pre-print platforms, and therefore some papers are not peer-reviewed. CSL is collected from published journal papers and is potentially of higher quality. In addition, CSL directly accesses the database without PDF/LaTeX parsing, which has near-perfect accuracy.

\subsection{Evaluation Benchmark}

The CSL contains meta-information provided by authors when submitting their papers, and the connections between them can constitute many NLP tasks.
In this section, we build a benchmark to facilitate the development of Chinese scientific literature NLP.
It contains diverse tasks, ranging from classification to text generation, representing many practical scenarios. 
We randomly select 10,000 samples and split the datasets into training set, validation set and test set according to the ratio, 0.8 : 0.1 : 0.1. 
This split is shared across different tasks, which allows multi-task training and evaluation.
From CSL, many possible combinations can also constitute tasks. We provide a toolkit for users to design tasks by their needs.

\textbf{Text Summarization (TS)} The paper title can be seen as a summary of the paper abstract. We build a summarization task predicting the paper title from the abstract. Existing Chinese text summarization resources are mainly concentrated in the news domain \cite{hu2015lcsts, liu2020clts}, and we provide the first text summarization task in the academic domain.

\textbf{Keyword Generation (KG)} In this task, the model is asked to predict a list of keywords from a given paper title and abstract. This task is similar to the Paper Topic Classification \cite{cohan2020specter}, but instead of predicting topics in a set of candidates, the goal is to generate keywords that correspond to the paper. We construct a dataset of paper's keywords, title and abstract. In English, there are related datasets such as SemEval \cite{kim2013automatic} and KP20k \cite{meng2017deep}. To the best of our knowledge, CSL provides the first Chinese keyword generation task.

\textbf{Text Classification} This task is predicting the category and discipline based on other information about the paper. We build a dataset for \textbf{category classification (CTG$_{\textup{cls}}$)}, which predicts the category with the paper title. Besides, we build a \textbf{discipline classification (DCP$_{\textup{cls}}$)} task that predicts the discipline with the paper abstract.

\section{Experiments}

\subsection{Baseline Models}

For baselines, we evaluate multi-task learning models trained on CSL tasks.
We use the text-to-text (i.e., feed text to produce text) method to unify downstream tasks in different formats. Specifically, these tasks are represented as the language generation task guided by a textual prompt. 
%For example, the target of keyword generation is to generate a sequence of keywords connected by special tokens (like ``word$_1$\_word$_2$\_word$_3$'').
We adopt several widely used text-to-text models, including T5 \cite{raffel2020exploring}, PEGASUS \cite{zhang2020pegasus}, and BART \cite{lewis2019bart}.
Since there are few publicly available versions of them, we conduct pre-training on the Chinese corpus from scratch.
In addition, we train a T5 using CSL paper abstracts as the corpus, namely CSL-T5, to provide a pre-training model that adapts to the Chinese scientific domain.

%For evaluation metrics, we adopt accuracy for classification tasks.
%We use ROUGE-L for the summarization task, which is commonly used for language generation tasks. For keyword generation, we use Bpref., which evaluates both the accuracy and order of generated keywords.
%All the metrics are calculated at Chinese character level.

\subsection{Settings}

For pre-training Chinese text-to-text models, we follow the architecture, optimization, and hyperparameter choices described in the papers. Following Google's Chinese BERT \citep{devlin2019bert}, we use the tokenizer with a vocabulary of 21,128 Chinese characters.
Models are pre-trained on the CLUE Corpus Small \citep{xu2020cluecorpus2020} for 1M steps with the batch size of 512.
We progressively train CSL-T5 basis on pre-trained T5, using the paper abstract as the corpus for 20k steps with the same hyperparameters.

Experiments are conducted on UER-py framework \cite{zhao2019uer} \footnote{https://github.com/dbiir/UER-py}.The learning rate is set to $3e^{-4}$ for T5; $1e^{-5}$ for BART and PEGASUS. The batch size is 32.
For multi-task training, we combine the training sets of each task for training 5 epochs. We use a prompt to specify which task the model should perform, e.g., ``to category'' for category classification. Then, we fine-tune the models on the task to be evaluated for 3 epochs with early stopping. All results are reported with greedy decoding.

\begin{table}[h]
\center
\small
\begin{tabular}{lcccc}
\toprule
\multirow{2}{*}{Models} & CTG$_{\textup{cls}}$ & DCP$_{\textup{cls}}$ & TS & KG  \\
\cmidrule(lr){2-2} \cmidrule(lr){3-3} \cmidrule(lr){4-4} \cmidrule(lr){5-5}

 & Acc. & Acc. & R-L & Bpref.  \\
\midrule
\multicolumn{1}{l}{T5} & \textbf{83.6} & 67.1 & 49.8 & 54.2  \\
\multicolumn{1}{l}{PEGASUS} & 81.7 & 69.4 & 49.4 & 55.2  \\
\multicolumn{1}{l}{BART} & 79.2 & 65.7 & 47.8 & 49.9  \\
\multicolumn{1}{l}{T5 (single)} & 82.3 & 63.2 & 49.2 & 54.1  \\
\midrule
\multicolumn{1}{l}{CSL-T5} & 82.9 & \textbf{70.8} & \textbf{52.1} & \textbf{55.9}  \\
\bottomrule
\end{tabular}
\caption{\label{exp}
The test performances of baseline models on CSL downstream tasks. T5 (single) is the result of fine-tuning T5 on each task separately, and the remaining columns are the results of multi-task learning.
}
%\vspace{-0.5cm}
\end{table}

\begin{CJK*}{UTF8}{gbsn}

\begin{table}[!htbp]
\linespread{1.16}
\small
\begin{tabular}{|m{7.1cm}|}
\hline
\textbf{Prompt:} to title   \\

\textbf{Input Text:} 通过对美国职业排球运动员进行非结构性访谈研究美国职业排球运动员对赞助商和赞助行为的态度 ... 赞助商应尊重运动员的情感和观点,从而使双方都能获得长远利益. \\
Through interviews, research was conducted on the attitudes of American professional volleyball players regarding sponsors and sponsorship activities ... Sponsors should respect athletes' feelings and opinions in order for both sides to profit in the long run. \\
\textbf{Prediction:} 美国职业排球运动员对赞助商和赞助行为的态度研究  \\
Research on American Professional Volleyball Players' Attitudes Towards Sponsors and Sponsorship Behaviors \\
\textbf{Ground Truth:} 美国排球运动员对赞助的态度分析   \\ 
Analysis of American Volleyball Players' Attitudes towards Sponsorship \\
\hline

\textbf{Prompt:} to keywords   \\

\textbf{Input Text:} 通过对祁连山自然保护区周边农牧民经济状况的调查发现阻碍经济发展的问题 ... 提出了发展生态旅游等适合本地区经济发展的模式. \\
Problems with economic development were discovered during an investigation of the economic conditions of farmers and herders in the Qilian Mountain Nature Reserve ... Ecotourism and other models for local economic development were proposed. \\
\textbf{Prediction:} 祁连山自然保护区; 农牧民; 经济发展模式  \\
Qilian Mountain Nature Reserve; Peasants and herdsmen; Economic development model\\
\textbf{Ground Truth:} 祁连山自然保护区; 周边经济; 发展模式  \\ 
Qilian Mountain Nature Reserve; Peripheral economy; Development model \\
\hline
\end{tabular}
\caption{\label{samples}
Samples of text summarization and keyword generation of CSL-T5.
}
\end{table}

\end{CJK*}

\subsection{Overall Performance}

The experimental results are shown in Table~\ref{exp}. Output samples of text summarization and keyword generation tasks are shown in Table~\ref{samples}.
For text classification, we report accuracy.
We use ROUGE-L \cite{lin2003automatic} for the summarization task, which is commonly used for language generation tasks. 
For keyword generation, we use Bpref. \citep{buckley2004retrieval}, which evaluates both the accuracy and order of generated keywords.
We can observe that baseline models can achieve acceptable results, where T5 outperforms other models. However, it is still not satisfactory for real-world applications, and future efforts are needed.
We also find that domain-adaptive training can further improve T5's performance. Similar experiments are also done by \citet{beltagy2019scibert} and \citet{gururangan2020don}, it partially demonstrates the value of CSL corpus for pre-training. The model and corpus will be publicly available.

%The BART is relatively insufficient on the keyword generation task. One of the possible reasons is that this task asks the model to generate a set of keywords connected by special tokens (like ``word$_1$\_word$_2$\_word$_3$''), while BART is pre-trained with language modeling objective. In contrast, the pre-training targets of T5 and PEGASUS are consist of discrete paragraphs (tokens), which may be closer to the keyword generation target.

%(1) T5 can outperform other baseline models with a comparable amount of training and parameters, with an average score of 63.6. However, it is still not satisfactory for real-world applications.
%(2) In classification tasks, DCP$_{\textup{cls}}$ is more difficult than CTG$_{\textup{cls}}$. This indicates that even given more informative input (paper abstract), it is challenging to distinguish between numerous disciplines. Therefore, it still needs exploration of the classification methods of Chinese scientific literature.
%(3) Existing methods are relatively inefficient for summarization and keyword generation tasks. It suggests that it is necessary to design generative pre-training models for scientific domain.
%(4) Domain-adaptive pre-training can improve T5's performance by 1.5\% on average. Similar experiments are also done by \citet{beltagy2019scibert} and \citet{gururangan2020don}, it partially demonstrates the value of CSL corpus for pre-training.

To discover the effect of multi-task training, we fine-tune T5 with each task individually.
From the comparison between T5 and T5-single, multi-task learning slightly outperforms individually fine-tuned models. We speculate that since the CSL tasks are homogeneous (derived from the same dataset), it is easier for models to share knowledge across different tasks.

CSL can create a large number of tasks by different combinations of tasks' input and output. It provides a natural playground for observing which tasks are mutually reinforcing when learned together. CSL could also be useful for cross-task research \cite{ye2021crossfit, bragg2021flex}. For example, exploring which tasks the model learns can help it quickly adapt to new tasks. We leave that for future exploration.

\section{Conclusion}

This paper presents the first Chinese scientific literature dataset, CSL, which can serve as a pre-training corpus and can derive abundant NLP tasks. Based on CSL, we build an evaluation benchmark to explore the challenges posed by automatic analysis of Chinese scientific documents.
Experimental results find difficulties in existing models in the Chinese scientific domain and point out the future directions. 

\textbf{Limitations and future work.} In the current version of CSL, to provide accurate category/discipline labels, we only use journals focused on one domain, which resulted in some data loss.
In future work, we will provide multi-label datasets to cover cross domain papers, and annotate CSL with more attributes like Chinese-English parallel data for academic machine translation.
Also, the versatile NLP task derived from CSL constitutes a naturally cross-task scenario. In the future, we will explore the role of CSL in cross-task and few-shot research.

%In the future, we will explore the role of CSL in cross-task research, and annotate CSL with more attributes like Chinese-English parallel data for academic machine translation.
% Entries for the entire Anthology, followed by custom entries

\section*{Ethical Considerations}

The corpus we use is released by the Chinese government aimed at sharing academic resources, which has been anonymized wherever necessary. We are licensed to use some of paper's metadata for NLP research. Therefore, our dataset does not involve any privacy or copyright issues.

\section*{Acknowledgments}

This research was supported by National Natural Science Foundation of China under grant no. 91959108 and Guangdong Basic and Applied Basic Research Foundation under Grant 2020A1515111199. 
We appreciate the data preprocessing and annotation effort done by Zichun Cao, Peixia Zhang, and Zhihui Wu in Prof. Zhang's lab at CUGB.

\bibliography{anthology,custom}
\bibliographystyle{acl_natbib}

%\appendix

%\section{Example Appendix}
%\label{sec:appendix}

%This is an appendix.

\end{document}